\documentclass[sigconf]{acmart}

\AtBeginDocument{%
  }

\setcopyright{acmlicensed}
\copyrightyear{2026}
\acmYear{2026}
\acmDOI{XXXXXXX.XXXXXXX}
\acmConference[WWW '26]{Industry Tracks - The Web Conference 2026}{April 13--17,
  2026}{Dubai, UAE}
\acmISBN{978-1-4503-XXXX-X/2026/06}

\begin{document}

\title{Getting Your Indices in a Row:\\ Full-Text Search for LLM Training Data for Real World}


\author{In\'es Altemir Mari\~{n}as}
\affiliation{%
  \institution{IC, EPFL}
  \city{Lausanne}
  \country{Switzerland}}
\email{<first.last>@epfl.edu}

\author{Anastasiia Kucherenko}
\affiliation{%
  \institution{IEM, HES-SO Valais-Wallis}
  \city{Sierre}
  \country{Switzerland}
}
\email{<first.last>@hevs.ch}

\author{Alexander Sternfeld}
\affiliation{%
  \institution{IEM, HES-SO Valais-Wallis}
  \city{Sierre}
  \country{Switzerland}
}
\email{<first.last>@hevs.ch}

\author{Andrei Kucharavy}
\affiliation{%
  \institution{II, HES-SO Valais-Wallis}
  \city{Sierre}
  \country{Switzerland}
}
\email{<first.last>@hevs.ch}

\renewcommand{\shortauthors}{Altemir Mari\~{n}as et al.}

\begin{abstract}

The performance of Large Language Models (LLMs) is determined by their training data. Despite the proliferation of open-weight LLMs, access to LLM training data has remained limited. Even for fully open LLMs, the scale of the data makes it all but inscrutable to the general scientific community, despite potentially containing critical data scraped from the internet.

In this paper, we present the full-text indexing pipeline for the Apertus LLM training data. Leveraging Elasticsearch parallel indices and the Alps infrastructure—a state-of-the-art, highly energy-efficient arm64 supercluster —we were able to index 8.6~T tokens out of 15.2~T used to train the Apertus LLM family, creating both a critical LLM safety tool and effectively an offline, curated, open web search engine. Our contribution is threefold. 
First, we demonstrate that Elasticsearch can be successfully ported onto next-generation arm64-based infrastructure.
Second, we demonstrate that full-text indexing at the scale of modern LLM training datasets and the entire open web is feasible and accessible.
Finally, we demonstrate that such indices can be used to ensure previously inaccessible jailbreak-agnostic LLM safety. 

We hope that our findings will be useful to other teams attempting large-scale data indexing and facilitate the general transition towards greener computation.




\end{abstract}



\maketitle

\section{Introduction}

Modern Large Language Models (LLMs) are becoming increasingly integrated in products and services. Based on a simple principle - pretraining for next or missing tokens prediction~\cite{ELMo2018AI2, GPT2018OpenAI} - followed by universal instruction-following fine-tuning~\cite{FLAN2022Google, InstructGPT2022OpenAI}, their impressive multidomain capabilities stem primarily from pretraining on massive web-scale datasets containing elements of texts required for instruction-following~\cite{TextbooksIsAllYouNeed2023}. 
While most LLM training datasets are closed, several ubiquitous large-scale public datasets emerged, notably  \textit{Common Crawl}, and its derivatives, such as \textit{The Pile} ~\cite{gao2020pile800gbdatasetdiverse},\textit{C4} ~\cite{JMLR:v21:20-074}, \textit{FineWeb} ~\cite{penedo2024finewebdatasetsdecantingweb}, or \textit{FineWeb2}~\cite{fineweb2}, among many others. Common Crawl and its derivatives contribute to the training of most widely used LLMs, such as GPT-3, for which it supplied 80\% of the data~\cite{brown2020languagemodelsfewshotlearners}.


However, the training data scale is a double-edged sword. The same scale that enables the emerging LLM multidomain capabilities also enables highly problematic emerging behaviors~\cite{AIBorder2022Anthropic}. The web-mined corpora, often not publicly shared, inevitably contain problematic content leading to problematic LLM behaviors~\cite{luccioni, mendu2025saferpretraininganalyzingfiltering}, that must be carefully analyzed to understand model capabilities and risks~\cite{10.1145/3458723, 10.1145/3287560.3287596}.

Beyond the unprompted LLM toxicity arising from the inclusion of the baseline social media toxicity into the crawls~\cite{mcguffie2020radicalizationrisksgpt3advanced, gehman-etal-2020-realtoxicityprompts}, training data has been discovered to contain private, copyrighted, abusive, or intentionally misleading information, among others. 
Personal information, such as names, social security numbers, and contacts, leaking from training data has been successfully extracted from GPT-2 and ChatGPT~\cite{carlini2021extractingtrainingdatalarge,TramerNewExtraction2023}, following their extensive memorization abilities from single examples ~\cite{10.5555/3495724.3495883, Carlini2022QuantifyingMA}. 
New York Times has blocked OpenAI’s crawler and sued them over unauthorized copyrighted content use for model training~\cite{NYTvsOpenAI2024}. 
Child Sexual Abuse Material (CSAM) has been discovered in the training data of several models~\cite{ciardha2025aigeneratedchildsexual}. Malicious actors have been discovered to inject disinformation into LLM training and reference data, such as Kremlin-affiliated Pravda Network~\cite{newsguard2025moscowAI}.


Problematic training data leads to problematic LLM behaviors. Even for API-only models with generation-time filtering and guardrails, jailbreaking and imperfect guardrails have led to highly problematic content generation, resulting in disastrous real-world impacts~\cite{cnbc2025palmSpringsAI}. For open-weight models, such protection is impossible, and not only can they be easily jailbroken~\cite{mazeika2024harmbenchstandardizedevaluationframework}, especially in a multilingual setting~\cite{AllLanguagesSecurity2024}, but any alignment tuning can be removed through post-training~\cite{R11776Perplexity2025}, unlocking problematic capabilities acquired from the training data. As such, the only reliable approach to LLM security and safety is in-depth analysis and curation of LLM training data \cite{obrien2025deepignorancefilteringpretraining}. Yet, in practice, even for open-weight models, systematic examination of training datasets remains extremely challenging. Widely used models typically disclose only broad information about their data sources, limiting transparency, reproducibility, and ultimately our ability to assess or guarantee their safety. 

As part of the Apertus LLM~\cite{apertus} transparency effort, we address this issue a single LLM family at a time, by indexing for full-text search the already open and reproducible Apertus LLM training data. To achieve it, we scale up a prior approach based on Elasticsearch full-text indexing allowing for fuzzy matching and logic operations~\cite{elazar2024whatsbigdata}, leveraging the Alps supercluster Machine Learning Platform, built and managed by the Swiss National Supercomputing Centre (CSCS)~\cite{Alps2025CSCS, AlpsML2025}.

Our contributions are threefold:
\begin{itemize}
\item \emph{Elasticsearch on ARM64:} We provide a technical overview of challenges, solutions, and configuration overviews for deploying Elasticsearch on ALPS, an ARM64-based GH200 high-performance computer (HPC) supercluster. While ARM64 offers future-ready energy efficiency, it poses compatibility challenges even for the most common software developed and deployed on x86/amd64 systems. 
\item \emph{Indexing and performance evaluation:} We index an 8.6~T token dataset of the 15.2~T Apertus training and pretraining dataset, doubling the size compared to the previous largest full-text index of that type, Infinigram~\cite{infinigram}, and quadrupling the previous Elasticsearch-based one, WhatIsInMyBigData~\cite{elazar2024whatsbigdata}.
We open-source our code~\footnote{\href{https://github.com/Reliable-Information-Lab-HEVS/apertus-pretraining-data-indexing}{https://github.com/Reliable-Information-Lab-HEVS/apertus-pretraining-data-indexing}} and report the performance during indexing and search for a range of deployment configurations, allowing for replication and future work on web-scale data indexing.
\item \emph{Security and safety use cases:} We demonstrate two LLM safety and security usecases of LLM full-text indices, focusing on derogotary language in a multilingual setting and helpfulness for indiscriminate weapons creation.
\end{itemize}

By shifting attention from post-training alignment to the foundational quality of training data, our approach provides the scalable infrastructure needed for safer, more transparent, and more accountable AI development.

\section{Related Work}

The main technical challenge in creating pipelines for transparent auditing of LLM training data is the massive size of these datasets. As a result, many existing tools attempt to overcome these limitations in various ways. Some methods perform approximate membership inference to reduce computational costs, such as Data Portraits~\cite {marone2023dataportraits}. Others analyze only small statistical samples—for example, 1\% of Common Crawl (circa 81~GB), such as \citeauthor{luccioni} in~\cite{luccioni}, or a one-million-random-web-pages sample in~\cite{mendu2025saferpretraininganalyzingfiltering}. While such approaches are excellent for exploratory safety work, guaranteeing LLM safety and security requires full training dataset coverage.


Several studies approached smaller datasets. Tools such as Data Measurements on HuggingFace ~\cite{luccioni2021datameasurements} and Know Your Data from Google~\cite{google2021knowyourdata} were created to automate and improve data documentation for smaller, specific datasets. Additionally, in-depth analyses have been performed on specialized datasets, such as the COVID-19 Open Research Dataset \cite{zhang-etal-2020-covidex}, the QuoteBank corpus \cite{10.1145/3477495.3531696}, and medical data \cite{niezni2023extending}. However, none of them have been developed for the scale the modern SotA LLM training requires.

World-Wide-Web scale data indexing for LLM safety and security has gained increasing interest in recent years. For example, \citeauthor{dodge2021documentinglargewebtextcorpora} in~\cite{dodge2021documentinglargewebtextcorpora} analyzed multiple problematic aspects of the C4 corpus(156~B tokens, English-only) and created an Elasticsearch full-text index for it.
\citeauthor{piktus-etal-2023-roots} in~\cite{piktus-etal-2023-roots} were the first to offer both fuzzy and exact search on ROOTS, a 1.6~TB multilingual text corpus. The WhatIsInMyBigData tool~\cite{elazar2024whatsbigdata} applied ElasticSearch to several datasets, including C4 (156~B Tokens),  RedPajama (1.4~T tokens), and The Pile (380~B tokens). While laudable initiatives, covering full training data of older LLMs, such as Eleuther GPT, BLOOM, or Pythia families~\cite{BLOOMZnMT02022, GPTNeoX20B2022EleutherAI, PythiaFamily2023}, these approaches have not scaled to the SotA multi-trillion token territory. 

Finally, most recently, Infinigram~\cite{infinigram} has used Burrow-Wheeler Transform-like suffix arrays, enabling millisecond-fast exact search on datasets such as OLMo 2 Instruct (4.6~T tokens), Dolma-v1.7 (2.6~T tokens), RedPajama, and The Pile. While an impressive performance and arguably the only SotA LLM training dataset index, Infinigram only allows for exact text matching. In turn, this makes fuzzy search and logic operators significantly more involved, which is problematic given that numerous downstream tasks require them, such as training data attribution of LLM-generated texts~\cite{wuhrmann-etal-2025-low}.

\section{Data and Tools} 

We first describe the datasets that were indexed and the datasets of search queries used for analysis (Section~\ref{preliminaries:data}), then present Elasticsearch, the full-text search system employed for indexing and search (Section~\ref{preliminaries:elasticsearch}), and finally outline the technical details of the computing environment used in this work (Section~\ref{preliminaries:computing}). All the code used to perform the indexing and querying of the database is available  \href{https://github.com/Reliable-Information-Lab-HEVS/apertus-pretraining-data-indexing}{https://github.com/Reliable-Information-Lab-HEVS/apertus-pretraining-data-indexing}.

\subsection{Data}\label{preliminaries:data}

\subsubsection{FineWeb} \label{fineweb}
The FineWeb dataset~\cite{fineweb} is derived from Common Crawl, and serves as the foundation for large-scale multilingual web text pretraining. It aggregates web documents from a diverse set of sources, applying filtering and deduplication to ensure linguistic diversity and textual quality. In the training of Apertus, three additional datasets derived from FineWeb are used: FineWeb-Edu, FineWeb-2 and FineWeb-2-HQ~\cite{fineweb-edu, fineweb2, fineweb-2-hq}. FineWeb-Edu filters educational web pages from the FineWeb dataset. Additionally, an educational quality classifier based on \texttt{Llama3-70B-Instruct}~\cite{LLaMA3Herd2024} is used to score each page, for further filtering. For Apertus training only pages with an educational quality score of 2 or higher are considered. Last, FineWeb-2-HQ considers the top 10\% quality documents of FineWeb-2 for the top 20 highest resource languages, based on a classifier using XLM-RoBERTa embeddings, and re-hydrades the deduplicated dataset, adding up to 8 copies of abundant high-quality documents \cite{fineweb-2-hq}.

\subsubsection{Apertus training data}
Apertus pretraining data underwent additional processing before being used in training. First, copyrighted material remaining in the training datasets is identified and removed. Second, data from from websites that have opted out of crawling by AI-related crawlers as of January 2025 is removed as well. Third, all detected email addresses, IP addresses and IBAN bank account numbers are replaced with anonymous markers. Finally, toxicity filtering is performed across nine languages, with custom-trained language-specific classifiers, with data for low-resource languages being preserved as is. The full pipeline for the creation of the training data has been made publicly available~\footnote{\url{https://github.com/swiss-ai/pretrain-data}} and more information on each of these filtering steps is described in the technical report of Apertus \cite{apertus}. 

\subsubsection{Indexed datasets} \label{pretraining data}
In this section, we describe the used subset of the Apertus training data. In total, we indexed a text volume of 8.6~T tokens, corresponding to 58\% of the current total training data (15~T), providing a dataset comparable in scale to the full training corpus. This enables the characterization of indexing and analysis efficiency on large-scale data. 

Specifically, we indexed all datasets included in Phase 1 of the Apertus pretraining, as described in the technical report~\cite{apertus}. While Phase 1 utilized approximately 5~T tokens, the datasets we indexed correspond to supersets that were also sampled in later phases, resulting in a higher Apertus training data coverage on our side. 
Table~\ref{tab:data_overview} displays the different components. The most prominent datasets are FineWeb-Edu~\cite{fineweb-edu}, the 33\% documents of highest quality from FineWeb-2-HQ~\cite{fineweb-2-hq}, and for the languages not included in FineWeb-2-HQ a randomly sampled 33\% from FineWeb-2~\cite{fineweb2}. Additionally, StarCoder code-specific dataset was included~\cite{StarCoder2023}, along with a Common Crawl math-focused subset, extracted as part of SmolLM2 LLM training~\cite{SMoLLM2025HuggingFace}. Finally, the permissively licensed books from Project Gutenberg~\cite{GutenbergDataset} data is added to the pretraining mix to evaluate the memorization of book-like content, while Poison data is performance-neutral data added as part of persistent pretraining data poisoning research~\cite{ApertusPoison2025}.


\begin{table}[H]
\begin{tabular}{@{}lll@{}}
\toprule
Dataset & & Tokens (B) \\ 
\midrule
FineWeb-Edu (Score-2) & & 4815 \\

\begin{tabular}[c]{@{}l@{}}FineWeb-2-HQ (33\% highest quality) and \\ \hspace{1em} FineWeb-2 (33\% sample of other languages)\end{tabular} 
& & 3557 \\

StarCoder & & 235 \\
FineMath CommonCrawl subset & & 32 \\
Gutenberg and Poison & & 2 \\ 
\bottomrule
\end{tabular}
\caption{Overview of the datasets indexed, along with the total number of tokens in each dataset.} 
\label{tab:data_overview}
\end{table}

The core strength of both Apertus and FineWeb-2 dataset is the attention to multilinguality. For this reason, we also focus our attention in Section~\ref{section:harm_analysis} on the presence of harmful content across various languages. Given that Apertus was developed in Switzerland, in addition to English we consider Swiss German (\verb|gsw|) and the official Swiss languages (Italian (\verb|ita|), German (\verb|deu|), French (\verb|fra|)) that are high-resource. Additionally, we attempt to sample non-european and lower-resource languages by focusing on Arabic, Thai, Kabyle, and Filipino. Finally, we investigate Esperanto as an example of synthetic language without a native speaker population.




\subsubsection{Datasets of search queries} \label{datasets_searchqueries}
One goal of the indexing and search pipeline is to examine the presence and distribution of potentially harmful or otherwise problematic content within the Apertus training data. For this, we focus on three LLM security and safety-related use cases:

\begin{itemize}
    \item \textbf{Weaponized Words dictionary}\footnote{\url{https://weaponizedword.org/}}: Is a large multilingual lexicon of curse words, slurs, and other forms of toxic or discriminatory language, maintained covering over 137 languages across 236 countries. Table \ref{tab:ldnoobw} shows the number of terms for each of the languages we consider: English, Italian, French and German. The lengths of the harmful terms range from 1 to 3 words. Note that we cannot publicly disclose this dataset, as access is restricted to research teams upon special request per the Weaponized Word Terms of Service.
    
    \item \textbf{Obscene Words list~\cite{shutterstock_badwords}}: The \textit{List of Dirty, Naughty, Obscene, and Otherwise Bad Words} (LDNOOBW) is a collection of profane and offensive terms used for cleaning of the Colossal Clean Crawled Corpus~\cite{dodge-etal-2021-documenting}. The strength of the dataset is that it contains  contains terms for 28 languages, including several low-resource languages such as Kabyle and Thai. The length of the profanities ranges from 1 to 5 words. In our analysis we consider the languages English, Italian, French, German, Arabic, Filipino, Esperanto, Kabyle and Thai. Table \ref{tab:ldnoobw} shows for each language the number of profanities that is included in the dataset.

    \item \textbf{Chemical weapons dataset}: A curated list of dangerous chemical agents selected from \textit{A Laboratory History of Chemical Warfare Agents} by Jared Ledgard \cite{ledgard2006laboratory}. Containing detailed synthesis instruction of chemical warfare agents, this book has only been freely available on the internet and only recently was de-indexed, with excerpts from it likely being included into LLM pretraining dataset, potentially making them helpful for indiscriminate weapons creation. We focus on several simple blood agents and their precursors to find potentially problematic contents in the internet, while leaving general-purpose chemistry knowledge intact.
    The dataset displayed in Table \ref{tab:chemicals} contains 17 terms, and we focus on several of the languages most likely to be used within Switzerland to look for this type of information, notably Dutch, Spanish, Serbo-Croatian, and Portuguese.
\end{itemize}

For each dataset, we report raw occurrence counts rather than terms prevalence, given that prior research suggests that knowledge acquisition and memorization are determined by information occurrence rather than prevalence~\cite{carlini2021extractingtrainingdatalarge}.

\begin{table}[H]
\centering
\begin{tabular}{@{}lll@{}}
\toprule
& \multicolumn{2}{c}{Number of terms} \\ 
\cmidrule(lr){2-3}
Language & LDNOOBW & Weaponized Words \\ \midrule
English   & 403 & 592 \\
Italian   & 168 & 21 \\
French    & 91 & 44 \\
German    & 66 & 31 \\
Arabic    & 38 & - \\
Esperanto & 37 & - \\
Thai      & 31 & - \\
Kabyle    & 22 & - \\
Filipino  & 14 & - \\ 
\bottomrule
\end{tabular}
\caption{Overview of the languages from the LDNOOBW and Weaponized Words datasets, showing the number of profanities for each language.}
\label{tab:ldnoobw}
\end{table}

\begin{table}[H]
\begin{tabular}{@{}l@{}}
\toprule
Chemical term          \\ \midrule
Chloropicrin           \\
Bromopicrin            \\
Diphenylchloroarsine   \\
Adamsite               \\
Hydrogen cyanide       \\
Cyanogen chloride      \\
Phosgene               \\
Sulfur mustard         \\
Methyldichloroarsine   \\
Acetone peroxide       \\
Triacetone triperoxide \\
Flash powder           \\
Potassium Perchlorate  \\
Potassium Chlorate     \\
Glycerine              \\
Nitric Acid            \\
Cyanogen Bromide       \\ \bottomrule
\end{tabular}
\caption{Manually curated list of dangerous chemical agents from \textit{A Laboratory History of Chemical Warfare Agents} \cite{ledgard2006laboratory}}
\label{tab:chemicals}
\end{table}

\subsection{Elasticsearch} \label{preliminaries:elasticsearch}
\subsubsection{General idea} Elasticsearch is a distributed search and analytics engine designed for efficiently handling large-scale text and structured data, commonly considered as a NoSQL database or a local search engine. At its core, it transforms raw information into an optimized index that allows users to perform fast and flexible queries, ranging from simple keyword lookups to complex semantic searches. This indexing-and-search model makes it especially well-suited for working with massive, heterogeneous datasets, such as collections of web-scraped documents, leading to its choice by several previous projects analyzing LLM training data.

In the described pipeline, raw data is stored as Parquet files and streamed into the indexing process to avoid overwhelming memory resources. 
Before being stored, the text undergoes multiple levels of processing via configurable analyzers, which generate different searchable representations — from normalized text suitable for semantic search, to exact forms that preserve original formatting and structure.

Scalability is achieved through Elasticsearch’s distributed architecture. Data is partitioned into shards and indexed in parallel, allowing multiple workers to process separate portions of the dataset simultaneously. Performance is further enhanced through techniques such as bulk indexing, adjustable batch sizes, and dynamic refresh intervals, which balance throughput and responsiveness during large-scale operations.

For datasets too large to fit on a single node, advanced cluster management strategies allow a scalable search cluster. Separate Elasticsearch instances can each index a subset of the data, and later be merged into a unified search space using remote \verb|_reindex| operations. This approach maintains data integrity while allowing the system to scale beyond the limitations of individual nodes.

\subsubsection{Parameter Tuning}
\label{sec:parameter-tuning}

The indexing operation relies on the \\ \verb|elasticsearch.helpers.parallel_bulk| function, which \\ distributes bulk requests across multiple threads to enable concurrent ingestion into Elasticsearch. Achieving good runtime and memory performance requires careful tuning of several interdependent parameters, as poor settings can easily create CPU under-utilization, memory exhaustion, or request bottlenecks.  

\textbf{Thread count} determines the number of worker threads handling bulk requests. Each thread maintains its own request queue, which increases memory consumption proportionally. The value should not exceed the number of available CPU cores. Higher thread counts improve concurrency but also risk memory pressure if chunk sizes are large.  

\textbf{Chunk size} specifies the number of documents in each bulk request. Small chunks create more frequent bulk submissions, increasing overhead from request construction and processing inside Elasticsearch. Oversized chunks, on the other hand, increase memory usage and the chance of timeouts. The maximum feasible chunk size is bounded by the ratio of \verb|max_chunk_size| to \verb|avg_doc_size|:
\begin{equation}
    chunk\_size \leq \frac{max\_chunk\_size}{avg\_doc\_size}
\end{equation}

\textbf{Max chunk bytes} controls the maximum payload size of a bulk request. Larger values reduce the relative cost of bulk request management inside Elasticsearch but require more RAM per thread to hold the request in memory.  

\textbf{Queue size} defines the buffer between the main thread (producing chunks) and worker threads (processing them). A larger queue helps absorb temporary imbalances between production and consumption, but also increases memory footprint. In practice, values between 2 and 8 were tested.  

Parameter choices must therefore balance CPU parallelism, request management overhead, and memory constraints. Estimating memory cost from \verb|avg_doc_size| and bounding chunk size via \verb|max_chunk_size| provides a practical starting point, after which empirical tuning is essential to reach stable throughput without exceeding hardware limits.  
\textbf{Hardware considerations} must also be taken in account, potentially providing an upper limit on document insertion speed. While Elasticsearch uses a two-phase atomic commit protocol for single document insertion it does not natively support bulk atomic transactions\footnote{\url{https://www.elastic.co/blog/found-elasticsearch-as-nosql}}, meaning that in a multi-threaded environment, each document insertion required two sequential round-trips to the storage for each inserted document. On well-optimized storage infrastructure such as ALPS IOPStore we used to indexing, this drive access latency can approach 50 microseconds, with expected maximal indexing speeds around 10 000 documents per second. This limit is determined by hardware latencies rather than document size or indexing settings.


\subsubsection{Types of Queries} \label{querytypes}

With Elasticsearch, one has the possibility to perform various types of queries. For instance, one can perform exact queries, fuzzy queries, or perform boolean logic with manually configured boolean queries. In this work, we focus on the match phrase query, which is described in more detail below. For this query, the system processes content using the \\ \verb|web_content_analyzer|, which performs:

\begin{itemize}
    \item HTML stripping to remove tags
    \item Standard tokenization to split text into words
    \item Lowercase normalization
    \item ASCII folding to convert accented characters to plain letters
\end{itemize}


The \texttt{match\_phrase\_query} searches for an exact sequence of words, with a configurable tolerance for how closely those words must appear together. This tolerance is controlled by the \\ \texttt{MATCH\_PHRASE\_SLOP} parameter. For single-word queries, it behaves the same as a regular match query. For multi-word phrases such as ``climate change,'' the terms must appear in the same order after analysis. When the slop value is set to 0, the words must occur directly next to each other. For example, ``climat'' must immediately precede ``chang.'' Increasing the slop allows a greater distance between words: a slop of 1 permits one intervening word, and a slop of 2 allows two. In practice, this means a slop of 1 can match ``climate and change,'' while a slop of 2 can match ``climate action and change.''



\subsection{Computing Environment and Resource Usage}\label{preliminaries:computing}
\subsubsection{Computing environment}
The experiments were conducted on the Alps Research Infrastructure at the Swiss National Supercomputing Centre (CSCS). The large-scale high-performance computing (HPC) system is an HPE Cray EX system with a HPL performance of 434 PFlops, making it one of the most powerful existing supercomputers, sitting as the N 8 of Top 500~\footnote{\url{https://www.top500.org/lists/top500/list/2025/06/}}.

The system is equipped with Arm64-based NVIDIA Grace Hopper GH200 nodes. Each node combines Grace CPUs with integrated Hopper GPUs, providing four Grace-Hopper modules and associated network interfaces. In total, the system contains 10,752 nodes, with each node containing 4 GPUs and 4 CPU sockets. The storage includes a 100PB ClusterStor HDD system and a 3PB ClusterStor SSD system, alongside a 1PB VAST storage system. Sharing the architecture with the current top 1 system on the Green Top 500 list, Alps itself is highly energy-efficient, achieving 61 GFlops/watt and ranking 19th on that list\footnote{\url{https://www.top500.org/lists/green500/list/2025/06/}}.

Architecturally, Alps uses a software-defined cluster (vCluster), abstracting infrastructure, service management and user environments, bridging the gap between traditional HPCs and Cloud service provider deployments. Specifically, the ML-dedicated vCluster, Clariden, is a container-first cluster using Slurm as scheduler~\cite{AlpsML2025} Elasticsearch version 7.17.28 was deployed in a distributed setup across multiple nodes.


\subsubsection{Usage and emissions}
We give an estimation of the
$\text{CO}_2$ emitted by the computation. In total, we used 7741 node hours for indexing and search presented here, including trial runs and parameter tuning, totaling less than 0.1\% of the computational power used to train the Apertus Model. We assume a power usage of 560W per node, consistent with Apertus training load, resulting in 4.3 MWh for the development and execution of the data indexing. While Alps' electric supply is carbon-neutral~\cite{cscs_energy}, by performing a consumption substitution analysis, we estimate that $\text{CO}_2$ emissions per kilowatt-hour (kWh) in Switzerland are on average 21g $\text{CO}_2\text{eq}$/kWh, our work led to approximately 90kg $\text{CO}_2\text{eq}$, or around 0.225\% of the yearly emissions of an average Swiss person~\cite{yearly_emission, energy_usage}.



\begin{table*}[t]
\begin{tabular}{@{}lllllll@{}}
\toprule
                  &                & Data size (GB) & Time (h) & Indexing rate (doc/s) & Index size/data size & Avg. peak memory (GB) \\ \midrule
Fineweb-2 Edu (EN) & Score 2       & 12,736.9 & 143.7 & 10,296.4 & 1.3 & 4.9 \\
Fineweb-2 Europe  & High quality*   & 2,660.0 & 408.3 & 589.4 & 1.1 & 7.5 \\
                  & Medium quality & 21 & 3.2 & 1,932.5 & 2.2 & 5.5 \\
Fineweb-2 Other   & High quality   & 991.8 & 194.4 & 1,315.4 & 2.8 & 2.4 \\
                  & Medium quality & 76.6 & 0.8 & 5,868.3 & 2.3 & 8.0 \\
Finemath-3        &                & 47 & 0.4 & 11,062 & 1.7 & 7.4 \\
Starcoder         &                & 229.1 & 4.2 & 10,919.3 & 1.4 & 12.7 \\
Gutenberg         &                & 3.2 & 0.05 & 1,302.2 & 2.4 & 4.9 \\ 
Poison            &                & 0.6 & 0.05 & 2,410.3 & 1.6 & 2.4 \\ \bottomrule
\end{tabular}
\caption{Time required to index each part of the phase 1 training data, alongside the data size, indexing rate, overhead, and average peak memory. * Designates the deduplicated dataset index.}
\label{tab:indexing_performance}
\end{table*}

\section{Elasticsearch Deployment}

Deploying Elasticsearch on high-performance computing (HPC) systems equipped with Grace Hopper nodes and Arm64 architectures presents several nontrivial challenges that do not arise in more conventional x86-based environments. A central difficulty lies in containerization. Many HPC platforms provide their own container engines, optimized for workload scheduling and security, but incompatible with Docker. This incompatibility reflects both architectural and security concerns: the majority of Docker images target x86\_64/amd64 rather than Arm64, and Docker’s shared read/write permissions are unsuitable for multi-user HPC systems. As a result, the official Elasticsearch image and orchestration tools such as Docker Compose cannot be used.

To overcome these limitations, we built custom OCI-compliant container images using Podman\footnote{\url{https://podman.io/}}. These images bundled the appropriate version of Elasticsearch together with supporting tools such as Python3, pyarrow, and curl, while remaining fully compatible with the Arm64 architecture and the HPC system’s security model. Orchestration, normally simplified by Docker Compose, was re-implemented through SLURM job definitions, ensuring reproducibility and integration with the resource manager. 

In an HPC environment, system-level restrictions can give further challenges. In our case, environment definition files, typically used to propagate configuration variables into containers, were not reliably interpreted by Elasticsearch. We addressed this by redirecting all runtime parameters through explicit command-line arguments injected at container startup. Similarly, Elasticsearch’s reliance on memory mapping conflicted with the immutable kernel parameter \texttt{vm.max\_map\_count}, which was set too low on the cluster to satisfy bootstrap checks. We resolved this by disabling the memory mapping, enabling Elasticsearch to start successfully. However, the disadvantage is that due to the absence of memory mapping we observed a reduction in the I/O efficiency and a higher system memory usage.

Last, networking required careful reconfiguration. In this HPC environment, HTTP traffic was subject to proxy mediation, leading to unexpected failures when attempting to connect to Elasticsearch’s default endpoint on localhost:9200. We resolved this by explicitly bypassing the proxy for local traffic, binding all Elasticsearch interfaces strictly to 127.0.0.1, and disabling multi-node discovery to comply with SLURM’s job isolation model. These settings ensured stable single-node operation suitable for HPC workloads.

\section{Performance Statistics}
We start by considering the indexing and search performance of Elastic Search on the Phase 1 training data, as described in Section \ref{pretraining data}. To this end, we perform three analyses. First, we look at the indexing statistics for each of the datasets. Then, we consider the effect of the query length on the search time. 

\subsection{Indexing Performance}


Table \ref{tab:indexing_performance} displays the indexing statistics for each of the datasets described in Section \ref{pretraining data}. The indexing performance varies notably with the linguistic composition and content type. Indexing pure English text proceeds substantially faster than indexing a multilingual dataset comprising a variety of European languages. The throughput for English text reached 10,297 documents per second, whereas the multilingual dataset achieved only 589 documents per second. This difference likely arises from the higher linguistic and computational complexity of multilingual data. English text is comparatively uniform in structure and encoding, with straightforward tokenization, while multilingual corpora introduce additional processing overhead due to diverse character sets, diacritics, and language-specific normalization routines.

Memory usage measurements show that the peak memory consumption is considerably higher when indexing code than when indexing natural language text. Code contains more unique tokens, longer identifiers, and complex syntactic patterns, which require larger intermediate data structures and less redundancy. As a result, code indexing demands greater memory resources.

The index overhead also differs significantly between datasets. For pure English, the ratio between index size and raw data size is relatively low (1.3), whereas it significantly increases when multiple languages are included. This indicates that English text is more easily compressed and represented efficiently, while multilingual datasets require larger vocabularies and less compressible structures due to linguistic variability.

Given that Fineweb-2-HQ was partially rehydrated for duplicated content, we verified if content de-duplication could be leveraged to compress the index size and accelerate the indexing on Fineweb-2 Europe High quality only. For this, we pre-computed full document SHA-256 hashes and inserted only the first document with the matching SHA-256 hash. Overall, we observed that approximately 68\% was duplicates, with most replicated content containing up to 8 copies, consistent with Fineweb-2-HQ documentation. Rather than increase in indexing speed, we observed an indexing slow-down proportional to duplicated content prevalence (approximately 68\%) lost in Fineweb-2 Europe High vs non-deduplicated Fineweb-2 Europe Medium, cf Table~\ref{tab:indexing_performance}. We interpret it as Elasticsearch managing duplicates internally with minimal overhead.

\subsection{Query Length and Search Time}
\begin{figure*}[]
    \centering
    \includegraphics[width=0.65\linewidth]{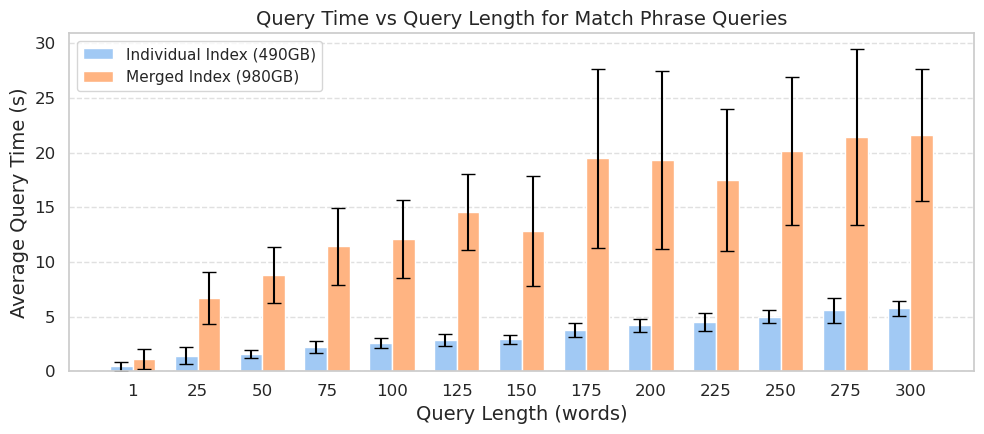}
    \caption{Query time vs query length for match phrase queries based on the index size. Averages +/- std.}
    \label{fig:querytime}
\end{figure*}

In Figure~\ref{fig:querytime} we evaluate the performance of match phrase queries when applied to an index with varying query segment length. To this end, for 13 different lengths between 1 and 300 words, we sampled 25 queries from the source datasets used to build the indexes. These segments were then issued as match phrase queries against the corresponding indexes to calculate the average search time and the standard deviation interval. We can observe a logical tendency of longer queries taking longer time, and get an intuition for overall performance.   


\section{Presence of Harmful terms in Apertus Pretraining Data} \label{section:harm_analysis}

\color{red}{Warning: This section contains samples of offensive words outside context.}\color{black}

In this section, we describe the analysis of the presence of harmful terms in the phase 1 training data of Apertus. To this end, we consider three Weaponized Words (WW), the manually constructed list of chemical warfare substances, and the LDNOOBW. Table \ref{tab:harmfulresult} shows the number of documents that contain at least one of the terms in each dataset. As expected, the results show that in low-resource languages there is fewer presence of harmful content. 


\begin{table}[H]
\begin{tabular}{@{}lccc@{}}
\toprule
               & \multicolumn{3}{c}{Count}                                                                                                                                                               \\ \cmidrule(l){2-4} 
Language       & \begin{tabular}[c]{@{}c@{}}WW\\ (x1,000,000)\end{tabular} & \begin{tabular}[c]{@{}c@{}}Chemical\\ (x1,000,000)\end{tabular} & \begin{tabular}[c]{@{}c@{}}LDNOOBW\\ (x1,000,000)\end{tabular} \\ \midrule
English        & 1,245.8                                                  & 2.96                                                         & 661.6                                                         \\
Italian        & 1.6                                                      & 0.23                                                          & 18.5                                                          \\
French         & 16.8                                                     & 0.12                                                          & 202.5                                                         \\
German         & 9.9                                                      & 0.58                                                          & 14.9                                                          \\
Dutch          & -                                                        & 0.12                                                          & -                                                             \\
Spanish        & -                                                        & 0.22                                                          & -                                                             \\
Serbo-Croatian & -                                                        & 0.04                                                          & -                                                             \\
Portuguese     & -                                                        & 0.22                                                          & -                                                             \\
Arabic         & -                                                        & -                                                            & 1.7                                                           \\
Filipino       & -                                                        & -                                                            & 0.6                                                           \\
Esperanto      & -                                                        & -                                                            & 3.1                                                           \\
Kabyle         & -                                                        & -                                                            & 0.0001                                                        \\
Thai           & -                                                        & -                                                            & 0.3                                                           \\ \bottomrule
\end{tabular}
\caption{Counts of document including terms deemed harmful according to: Weaponized Words (WW), Chemical substances and the LDNOOBW, Millions of hits.}
\label{tab:harmfulresult}
\end{table}

Furthermore, we observe that despite apparent high counts of problematic content, most highly represented words are general terms that may not always be problematic. These terms encompass general, important themes and cannot be universally removed from the training data without severely degrading models' performance, which is consistent with previous reports~\cite{dodge-etal-2021-documenting}. Driving this point further, despite the abundance of toxic words, Apertus performs well on toxicity evaluations~\cite{apertus} (Section 5.2, Table 26). However, full-text indexing enables fine-grained thematic analyses for granular further training data filtering and LLM behavior guarantees.

\subsection{Weaponized Words}
\begin{figure}
    \centering
    \includegraphics[width=1\linewidth]{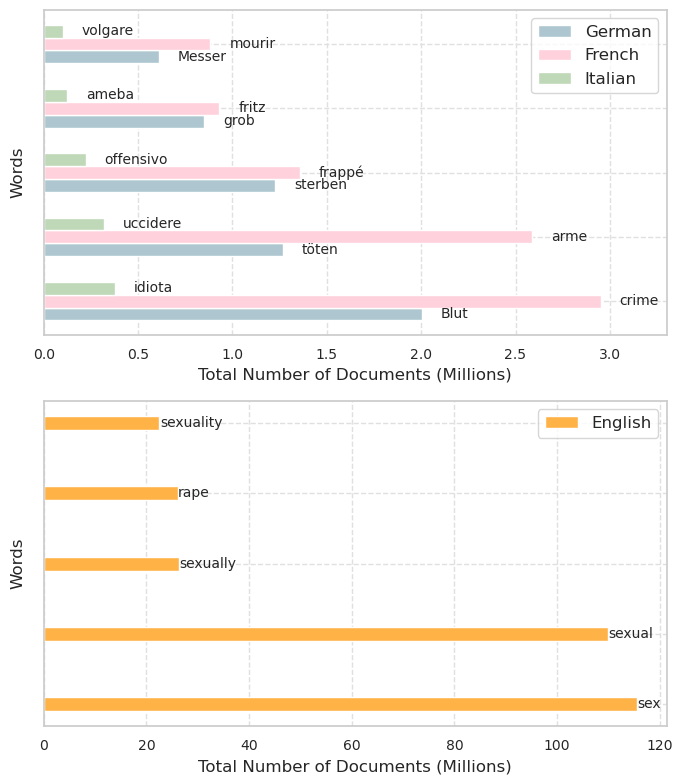}
    \caption{Top 5 Weaponized Words by total number of occurrences}
    \label{fig:ww_barchart}
\end{figure}

Figure \ref{fig:ww_barchart} shows, for each of the languages taken into consideration, the five most prominent terms in the training data. We observe that there are common terms across different languages, such as "dying" (\textit{mourir}, \textit{sterben}) and "killing" (\textit{uccidere, toten}). At the same time, there is a clear difference for the English language, in which terms related to \textit{sex} are the most common in the training data.

\subsection{Chemical Substances}
We observe that the counts are exceptionally high for common chemical compounds widely known to the general public, notably Glycerine, Nitric Acid, and Hydrogen Cyanide. In particular, most of their mentions would occur outside chemical weapons synthesis instructions. 
We observe that for rare terms unequivocally connected to chemical weapons synthesis (Bromopicrin, Methyldichloroarsine), substantial counts occur outside of English, indicating that multilingual data curation is essential to ensure model safety.

\begin{figure}[]
    \centering
    \includegraphics[width=\linewidth]{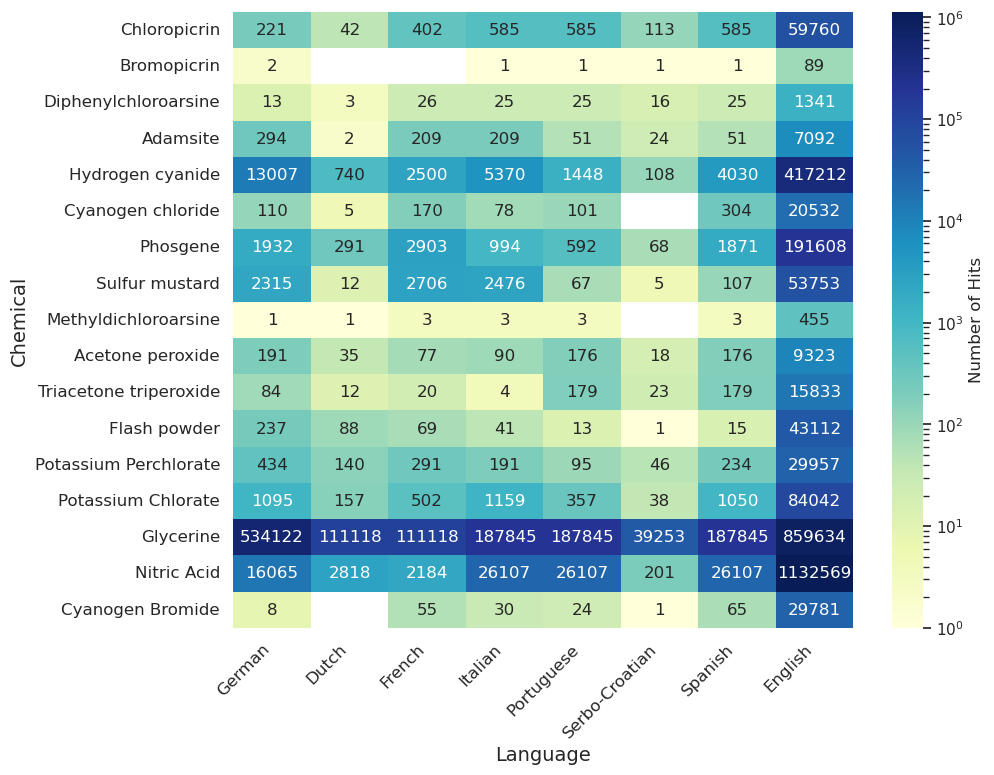}
    \caption{Heatmap displaying the presence of chemical terms in the training data for various languages.}
    \label{fig:heatmap}
\end{figure}


\begin{table}[]
\begin{tabular}{@{}lll@{}}
\toprule
Language  & Top 3 terms                 & \begin{tabular}[c]{@{}l@{}}Total count \\ (x 1,000,000)\end{tabular} \\ \midrule
English   & sex, sexual, sexually       & 251.9                                                                \\
French    & con, péter, bitte           & 189.6                                                                \\
German    & porno, penis, fick          & 6.6                                                                  \\
Filipino  & bobo, tanga, burat          & 0.5                                                                  \\
Kabyle    & qqu, abbuc, uqan            & 0.0001                                                               \\
Thai      & [Hī], [Yĕd], [Khwy]         & 0.2                                                                  \\
Esperanto & fik, piĉo, fek              & 3.0                                                                  \\
Arabic    & [faraj], [nik], [aghtisabi] & 0.4                                                                  \\ \bottomrule
\end{tabular}
\caption{Top 3 most found terms from the LDNOOBW for each of the languages taken into consideration, alongside the total count for those three terms. Enclosure by [] indicates transliteration.}
\label{tab:obscene}
\end{table}

Table \ref{tab:ldnoobw} displays the most common find terms from the LDNOOBW for each of the languages. Similar to Weaponized Words, we observe high counts of general words, which are necessary for discussing important topics.


\section{Conclusion and Future Work}

In this paper, we demonstrate how a full-text search index for 8.6~T multilingual tokens can be built on modern energy-efficient infrastructure and how it can be leveraged to ensure LLM behavior safety.
Our contribution is threefold. 
First, we demonstrate that, despite initial difficulties, common server-side applications -- such as Elasticsearch -- can be run on next-generation architectures. We believe this case study, which ported a common business application to an energy-efficient next-generation architecture, demonstrates a high readiness for a general green computing transition, critical as computational power demands are soaring. 
Second, we demonstrate that indexing texts at the scale of LLM training datasets is feasible for small teams using off-the-shelf software, requiring less than 0.1\% of the compute used to train the LLMs in question. We hope this will encourage future analyses of the LLM training data and the inclusion of full training data analysis in LLM release standards.
Finally, we demonstrate that the resulting index can be used for in-depth investigation of problematic content in the LLM training data, allowing for fine-grained context-based analysis and training data cleaning. We hope that this will lead to a common adoption of training data-based LLM safety and security mechanisms, such as those proposed by other teams, notably in~\cite{obrien2025deepignorancefilteringpretraining}.

While our training data indexing is primarily motivated by safety and security, the size of the indexed data begins to approach that of the Common Crawl dataset itself. In turn, this suggests that an offline search index for the entire open internet is within reach for smaller entities. An exciting perspective in itself, it is particularly interesting for the general-purpose factual rooting of LLM generation based on offline search curated open web subsets. In turn, this raises a host of ethical and economic questions, notably regarding the fair compensation of individuals who originally created and hosted the retrieved content, likely opening novel research directions in worldwide web economic, legal, and governance aspects.

\section*{Acknowledgements}

The Swiss Supercomputing Centre (CSCS) for providing computational support and infrastructure for this project; SwissAI initiative and the Apertus team for the insight regarding the training data and computational budget allocation; and Prof. Antoine Bosselut for co-supervising IAM during the duration of the project. This work has been supported by the armasuisse S+T grant AR-CYD-C-025 to AK, AK, and AS, and a SwissAI Initiative Large Project Grant 45.

\bibliographystyle{ACM-Reference-Format}
\bibliography{references}
\end{document}